# A Sliding Mode Force and Position Controller Synthesis for Series Elastic Actuators

Emre Sariyildiz, Rahim Mutlu, Haoyong Yu

*Abstract—* **This paper deals with the robust force and position control problems of Series Elastic Actuators (SEAs). It is shown that an SEA's force control problem can be described by a second-order dynamic model which suffers from only matched disturbances. However, the position control dynamics of an SEA is of fourth-order and includes matched and mismatched disturbances. In other words, an SEA's position control is more complicated than its force control, particularly when disturbances are considered. A novel robust motion controller is proposed for SEAs by using Disturbance Observer (DOb) and Sliding Mode Control (SMC). When the proposed robust motion controller is implemented, an SEA can precisely track desired trajectories and safely contact with an unknown and dynamic environment. The proposed motion controller does not require precise dynamic models of environments and SEAs. Therefore, it can be applied to many different advanced robotic systems such as compliant humanoids and exoskeletons. The validity of the motion controller is experimentally verified.**

## I. INTRODUCTION

In the last two decades, SEAs have received increasing attention due to their potential advantages (e.g., safety and high fidelity force control) in physical robot-environment interaction [1]–[3]. Several advanced robotic systems, such as compliant industrial, humanoid and exoskeleton robots, have recently been developed by using SEAs [4, 5]. A flexible mechanical component (e.g. spring) is intentionally placed between actuator and link in the design of an SEA [2, 3]. Although the compliant mechanical component improves physical robot-environment interaction, it complicates the mechanical design and motion control problem, particularly in position control [6, 7]. Moreover, it makes SEAs more sensitive to disturbances than inherently robust stiff actuators with high gear ratios. Therefore, the stability and performance of an SEA may significantly deteriorate by disturbances in practice [3, 7 and 8].

The position control problem of robots with flexible joints has long been studied in the literature. However, the proposed controllers have several shortcomings for SEAs. Singular Perturbation Method was applied to compliant robots in [9]. This method is sensitive to disturbances, and it can only be applied if the joint stiffness is high enough [9, 10]. Compliant robots have generally been controlled by using a PID-controller combined with a model-based feed-forward controller [11]–[13]. Although feed-forward control improves the positioning accuracy of links, it makes control system sensitive to disturbances. Moreover, it is applicable only for regulation control [13]. Intelligent control methods, such as neural network control, were applied to compliant robots in [14, 15]. They generally have high computational load and complex control structures to be implemented in real-time. Resonance Ratio Control (RRC) was originally proposed for the vibration control problem of industrial robots with flexible joints, and it was recently applied to SEAs in [7, 16]. An RRC-based position controller is sensitive to load and link uncertainties as the robustness is degraded to suppress the vibration of link [7]. Moreover, RRC cannot independently control whole system dynamics due to insufficient control parameters. Authors proposed the SMC-based robust position controller for SEAs in [17].

SEA studies have generally focused on force/impedance control problem and its advanced robotic applications, e.g., human-robot collaboration. A PID controller was proposed to perform force control in [2]. The performance of force control was improved by applying feed-forward control signal with a model-based controller. The proposed force control method suffers from stability, slow response and overshoot problems due to neglected disturbances such as backlash and stiction[3]. To tackle stability and performance problems, cascade force controllers (i.e. an inner-loop on motor velocity control and an outer-loop on force control) were proposed in [3, 18, 19]. However, the proposed cascade force controllers are sensitive to disturbances. A higher-order DOb was proposed for the inner-loop on motor velocity control so that the robustness of the force control system was improved in [20]. Later, DOb-based robust force controller was applied to the University of Texas' SEA (UT-SEA) and NASA's compliant humanoid robot Valkyrie [4, 21]. In classical control, a higher-order DOb synthesis is not straightforward, and it may significantly suffer from conservatism [22]. Bae et al. reported the robust force controller's unexpected stability problem and proposed an SMC force controller in [23]. To suppress the chattering of the SMC force controller, Wang et al. proposed a modified SMC controller by using nonlinear DOb and high gain control in [24]. Force control of an SEA was performed by using RRC in [7]. However, tuning the control parameters of RRC is not easy in force control. Authors recently proposed a new motion controller for SEAs in [25]. In this paper, an SEA's force control problem is described by a fourth-order dynamic model which suffers from matched and mismatched disturbances. Another high-precision robust force controller was recently designed in [26].

E. Sariyildiz and R. Mutlu are with the School of Mechanical, Materials, Mechatronic and Biomedical Engineering, University of Wollongong, Wollongong, Australia (phone:+61242213319; fax:+61242214577; e-mail: emre@uow.edu.au, rmutlu@uow.edu.au).

H. Yu is with the Depertment of Biomedical Engineering, National University of Singapore, Singapore. (e-mail: bieyhy@nus.edu.sg).

A novel motion controller is proposed for SEAs in this paper. The position and force controllers are synthesized by combining DOb and SMC. The former not only improves the robustness by canceling disturbances but also allows to reduce the control signal chattering. Two continuous motion controllers are proposed by using Quasi-SMC and Continuous-SMC. The main drawbacks of continuous SMC control methods are: the former degrades the robustness and the latter requires the estimation of acceleration. Continuous and discontinuous SMC methods are experimentally compared for SEAs in section V. An SEA's dynamic model is derived by considering its position and force/impedance control applications. It is shown that a second-order dynamic model, which suffers from only matched disturbances, can be used to represent the force control problem of an SEA. However, the dynamics of the position control of an SEA is of fourth-order and suffers from the mismatched disturbances in addition to matched disturbances. Therefore, an SEA's robust position control is more complicated than its robust force control. The proposed robust motion controller provides high-performance without precise dynamic models of an SEA and environment. For example, the link of an SEA can precisely track desired trajectories and stably contact with an unknown and dynamic environment. The proposed controllers can be applied to different compliant robotic systems driven by SEAs; e.g., exoskeletons, humanoids and collaborative industrial robots. Position and force control experiments are presented to validate the proposed sliding mode motion controllers.

The organization of the paper is as follows: an SEA's dynamic model is presented in section II; a second-order DOb is synthesized in section III; sliding mode motion controllers are synthesized in section IV; the controllers are experimentally verified in section V; and conclusion is given in section VI.

## II. Series Elastic Actuators

In this section, an SEA's dynamic model is derived for position and force/impedance control.

### A. Position Control:

If an SEA does not physically interact with an environment (See Fig. 1a), then its dynamic model is:

$$J_l^n \ddot{q} + b_l^n \dot{q} = \tau_s^n - \tau_l^d$$
$$J_m^n \ddot{\theta} + b_m^n \dot{\theta} = \tau_m - \tau_s^n - \tau_m^d \quad (1)$$
$$\tau_s^n = k^n (\theta - q)$$

where $J_m^n$ and $J_l^n$ are the nominal inertias of motor with speed reducer and link, respectively; $b_m^n$ and $b_l^n$ are nominal viscous friction coefficients; $k^n$ is the nominal spring constant of the SEA; $\theta$ and $q$ are the angles of motor and link sides, respectively; $\dot{\bullet}$ and $\ddot{\bullet}$ are the first and second order derivatives of $\bullet$ (i.e., the speed and acceleration of the motor and link sides), respectively; $\tau_m$ is the applied torque at motor side; $\tau_s^n$ is the nominal spring torque; $\tau_m^d$ is the matched disturbance; and $\tau_l^d$ is the mismatched disturbance.

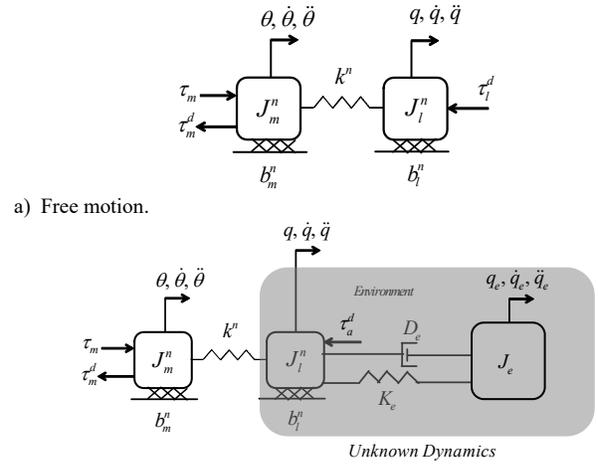

a) Free motion.

b) Physical SEA-environment interaction.
Fig.1: Dynamic models of an SEA.

The matched and mismatched disturbances of an SEA are:

$$\tau_m^d = \left(J_m - J_m^n\right)\ddot{\theta} + \left(b_m - b_m^n\right)\dot{\theta} + \tau_s - \tau_s^n + \tau_m^{ud} \quad (2)$$

$$\tau_l^d = \left(J_l - J_l^n\right)\ddot{q} + \left(b_l - b_l^n\right)\dot{q} + \tau_g + \tau_s^n - \tau_s + \tau_l^{ud} \quad (3)$$

where $J_m$ and $J_l$ are the uncertain inertias of motor with speed reducer and link, respectively; $b_m$ and $b_l$ are the uncertain viscous friction coefficients; $k$ is the spring constant of the SEA; $\tau_s = k(\theta - q)$ is the torque of the actuator's spring; $\tau_g$ is the gravitational disturbance; and $\tau_m^{ud}$ and $\tau_l^{ud}$ are unknown disturbances, respectively.

The position accuracy of an SEA's link may significantly deteriorate by external load and various internal disturbances, such as nonlinear frictions and hysteresis, due to its complex mechanical design [25]. To track the desired trajectory of an SEA's link with high-accuracy (e.g., $q = q^{des}$) $\tau_m^d$ and $\tau_l^d$ should be precisely suppressed/cancelled.

### B. Force/Impedance Control:

Let us describe force/impedance control goal of an SEA by using Hooke's law.

$$\tau_s^{des} = k(\theta - q)^{des} \quad (4)$$

where $\tau_s^{des}$ is the desired spring torque, and $(\theta - q)^{des}$ is the desired deflection of the actuator's spring [2].

With a simple mathematical manipulation and assuming that $q$ is known (i.e., the position of the SEA's link is measured), Eq. (4) can be rewritten as follows:

$$\theta^{des} = (1/k)\tau_s^{des} + q \quad (5)$$

where $\theta^{des}$ is the desired $\theta$.

Eq. (4) and Eq. (5) show that force/impedance control of an SEA can be conducted by precisely controlling its motor position/angle (i.e., $\theta = \theta^{des}$) [23, 25]. A precise force control can be performed by using acceleration feedback [27].

When the link of an SEA interacts with an environment, its dynamic model is derived from Fig. 1b as follows:

$$J_l^n \ddot{q} + b_l^n \dot{q} = \tau_s^n - \tau_l^d - \tau_{ext}^d \quad (6)$$

$$\tau_{ext} = \tau_a^d + J_e\left(\ddot{q} - \ddot{q}_e\right) + D_e\left(\dot{q} - \dot{q}_e\right) + K_e\left(q - q_e\right) \quad (7)$$

where $\tau_{ext}^d$ is the external disturbance torque; $\tau_a^d$ is the applied external torque; $J_e$, $D_e$ and $K_e$ are the inertia, damping and stiffness of the environment which SEA physically interacts, respectively; and $q_e$ is the angle of environment.

To obtain a practical model for an SEA which physically interacts with an unknown environment, let us assume that $\tau_l^d$ and $\tau_{ext}^d$ are unknown disturbances as shown in Fig. 1b. If Eq. (6) is substituted into Eq. (1), then Eq. (8) is derived as

$$J_m^n \ddot{\theta} + b_m^n \dot{\theta} = \tau_m - \tilde{\tau}_m^d \qquad (8)$$

where $\tilde{\tau}_m^d = \tau_m^d + J_l^n \ddot{q} + b_l^n \dot{q} + \tau_l^d + \tau_{ext}^d$.

Eq. (8) is a second-order dynamic model of a servo system which is disturbed by matched disturbances. In many advanced robot applications of SEAs, it is very hard to precisely model or measure the matched disturbances, e.g., the dynamics of human beings in human-robot interaction and rehabilitation. Such disturbances may significantly degrade the stability and performance of an SEA's force/impedance control if they are not treated in the controller synthesis [23]. To perform force/impedance control, i.e., to track the desired trajectory of the SEA's motor with high-accuracy $\left(\text{e.g.}, \theta = \theta^{des}\right)$, the matched disturbance $\left(\tilde{\tau}_m^d\right)$ should be precisely suppressed/cancelled.

## III. Second-order Disturbance Observer

In this section, a second-order DOb is synthesized so as to estimate an SEA's matched and mismatched disturbances and their first and second-order time derivatives [17, 25].

The state space representation of Eq. (1) is

$$\dot{\boldsymbol{\xi}} = \mathbf{A_n}\boldsymbol{\xi} + \mathbf{b_n}\tau_m - \boldsymbol{\tau_{dis}} \qquad (9)$$

where $\mathbf{A_n} = \begin{bmatrix} 0 & 1 & 0 & 0 \\ -k^n/J_l^n & -b_l^n/J_l^n & k^n/J_l^n & 0 \\ 0 & 0 & 0 & 1 \\ k^n/J_m^n & 0 & -k^n/J_m^n & -b_m^n/J_m^n \end{bmatrix}$, $\boldsymbol{\tau_{dis}} = \begin{bmatrix} 0 \\ \tau_l^d/J_l^n \\ 0 \\ \tau_m^d/J_m^n \end{bmatrix}$,

$\mathbf{b_n} = \begin{bmatrix} 0 & 0 & 0 & 1/J_m^n \end{bmatrix}^T$ and $\boldsymbol{\xi} = \begin{bmatrix} q & \dot{q} & \theta & \dot{\theta} \end{bmatrix}^T$.

The second-order DOb is synthesized by assuming that disturbance vectors are bounded, i.e., $\|\boldsymbol{\tau_{dis}}\| \leq \delta_{\tau_{dis}}, \|\dot{\boldsymbol{\tau}}_{dis}\| \leq \delta_{\dot{\tau}_{dis}}$, $\|\ddot{\boldsymbol{\tau}}_{dis}\| \leq \delta_{\ddot{\tau}_{dis}}$ and $\|\dddot{\boldsymbol{\tau}}_{dis}\| \leq \delta_{\dddot{\tau}_{dis}}$ where $\|\bullet\|$ is the norm of $\bullet$; $\dot{\boldsymbol{\tau}}_{dis}, \ddot{\boldsymbol{\tau}}_{dis}$ and $\dddot{\boldsymbol{\tau}}_{dis}$ are the first, second, and third-order derivatives of $\boldsymbol{\tau_{dis}}$, respectively; and $\delta_{\tau_{dis}} > 0$, $\delta_{\dot{\tau}_{dis}} > 0$, $\delta_{\ddot{\tau}_{dis}} > 0$ and $\delta_{\dddot{\tau}_{dis}} > 0$ are real numbers.

In order to estimate $\boldsymbol{\tau_{dis}}$, $\dot{\boldsymbol{\tau}}_{dis}$ and $\ddot{\boldsymbol{\tau}}_{dis}$, auxiliary variable vectors are defined by using

$$\mathbf{z_1} = \boldsymbol{\tau_{dis}} + L_1\boldsymbol{\xi} - \mathbf{z_2} \qquad (10)$$

$$\mathbf{z_2} = \dot{\boldsymbol{\tau}}_{dis} + L_2\boldsymbol{\xi} - \mathbf{z_3} \qquad (11)$$

$$\mathbf{z_3} = \ddot{\boldsymbol{\tau}}_{dis} + L_3\boldsymbol{\xi} \qquad (12)$$

where $\mathbf{z_j} \in \mathbb{R}^4$ is the $j^{th}$ auxiliary variable vector; and $L_j \in \mathbb{R}$ is the $j^{th}$ control gain of DOb, yet to be tuned.

Time derivatives of Eq. (10–12) are

$$\dot{\mathbf{z}}_1 = -\left(L_1 - L_2 + L_3\right)\left(\mathbf{z_1} + \mathbf{z_2}\right) + \mathbf{z_2} + \ddot{\boldsymbol{\tau}}_{dis} + \\ \left(L_1 - L_2 + L_3\right)\left(\mathbf{A_n}\boldsymbol{\xi} + \mathbf{b_n}\tau_m + L_1\boldsymbol{\xi}\right) - \left(L_2 - L_3\right)\boldsymbol{\xi} \qquad (13)$$

$$\dot{\mathbf{z}}_2 = -\left(L_2 - L_3\right)\left(\mathbf{z_1} + \mathbf{z_2}\right) + \mathbf{z_3} - \ddot{\boldsymbol{\tau}}_{dis} + \left(L_2 - L_3\right)\left(\mathbf{A_n}\boldsymbol{\xi} + \mathbf{b_n}\tau_m + L_1\boldsymbol{\xi}\right) - L_3\boldsymbol{\xi} \quad (14)$$

$$\dot{\mathbf{z}}_3 = -L_3\left(\mathbf{z_1} + \mathbf{z_2}\right) + \ddot{\boldsymbol{\tau}}_{dis} + L_3\left(\mathbf{A_n}\boldsymbol{\xi} + \mathbf{b_n}\tau_m + L_1\boldsymbol{\xi}\right) \qquad (15)$$

where $\dot{\mathbf{z}}_j \in \mathbb{R}^4$ is the time derivative of $\mathbf{z_j}$.

The dynamics of the auxiliary variable observer is:

$$\dot{\hat{\mathbf{z}}}_1 = -\left(L_1 - L_2 + L_3\right)\left(\hat{\mathbf{z}}_1 + \hat{\mathbf{z}}_2\right) + \hat{\mathbf{z}}_2 + \\ \left(L_1 - L_2 + L_3\right)\left(\mathbf{A_n}\boldsymbol{\xi} + \mathbf{b_n}\tau_m + L_1\boldsymbol{\xi}\right) - \left(L_2 - L_3\right)\boldsymbol{\xi} \qquad (16)$$

$$\dot{\hat{\mathbf{z}}}_2 = -\left(L_2 - L_3\right)\left(\hat{\mathbf{z}}_1 + \hat{\mathbf{z}}_2\right) + \hat{\mathbf{z}}_3 + \left(L_2 - L_3\right)\left(\mathbf{A_n}\boldsymbol{\xi} + \mathbf{b_n}\tau_m + L_1\boldsymbol{\xi}\right) - L_3\boldsymbol{\xi} \quad (17)$$

$$\dot{\hat{\mathbf{z}}}_3 = -L_3\left(\hat{\mathbf{z}}_1 + \hat{\mathbf{z}}_2\right) + L_3\left(\mathbf{A_n}\boldsymbol{\xi} + \mathbf{b_n}\tau_m + L_1\boldsymbol{\xi}\right) \qquad (18)$$

where $\hat{\bullet}$ is the estimation of $\bullet$. The second-order observer is synthesized by neglecting the estimation of $\ddot{\boldsymbol{\tau}}_{dis}$, i.e., $\dot{\hat{\ddot{\boldsymbol{\tau}}}}_{dis} = \mathbf{0}$.

Let us describe auxiliary variable estimation errors by using $\mathbf{e_1} = \mathbf{z_1} - \hat{\mathbf{z}}_1$, $\mathbf{e_2} = \mathbf{z_2} - \hat{\mathbf{z}}_2$ and $\mathbf{e_3} = \mathbf{z_3} - \hat{\mathbf{z}}_3$. If Eq. (13–15) are subtracted from Eq. (16–18), then we obtain

$$\dot{\mathbf{e}}(t) = \boldsymbol{\Psi}\mathbf{e}(t) + \boldsymbol{\tau_D}(t) \qquad (19)$$

where $\mathbf{e}(t) = \begin{bmatrix} \mathbf{e_1} \\ \mathbf{e_2} \\ \mathbf{e_3} \end{bmatrix}$, $\boldsymbol{\Psi} = \begin{bmatrix} -\left(L_1 - L_2 + L_3\right)\mathbf{I_4} & -\left(L_1 - L_2 + L_3\right)\mathbf{I_4} + \mathbf{I_4} & \mathbf{0_4} \\ -\left(L_2 - L_3\right)\mathbf{I_4} & -\left(L_2 - L_3\right)\mathbf{I_4} & \mathbf{I_4} \\ -L_3\mathbf{I_4} & -L_3\mathbf{I_4} & \mathbf{0_4} \end{bmatrix}$,

$\boldsymbol{\tau_D}(t) = \begin{bmatrix} \ddot{\boldsymbol{\tau}}_{dis} & -\ddot{\boldsymbol{\tau}}_{dis} & \ddot{\boldsymbol{\tau}}_{dis} \end{bmatrix}^T$, $\mathbf{I_4}$ is a $4 \times 4$ identity matrix, and $\mathbf{0_4}$ is a $4 \times 4$ null matrix.

The eigenvalues of $\boldsymbol{\Psi}$ correspond to the bandwidth of the observer and are derived by solving:

$$\lambda\left(\boldsymbol{\Psi}\right) = \det\left(\lambda\mathbf{I_{12}} - \boldsymbol{\Psi}\right) = \left(\lambda^3 + L_1\lambda^2 + L_2\lambda + L_3\right)^4 = 0 \qquad (20)$$

where $\lambda \in \mathbb{C}$. The bandwidth of DOb is $g_{DOb}$ rad/s when its control gains are $L_1 = 3g_{DOb}, L_2 = 3g_{DOb}^2$ and $L_3 = g_{DOb}^3$.

One can easily adjust disturbance estimation dynamics by tuning the control gains of DOb, i.e., assigning the eigenvalues of $\boldsymbol{\Psi}$. If the observer gains are properly tuned so that $\boldsymbol{\Psi}$ is negative definite, then input to state stability of Eq. (19) is achieved. In other words, all estimation errors which start in a circular plane whose radius is $\|\mathbf{e}(t_0)\| + \lambda_{min}^{-1}(\boldsymbol{\Psi})\delta_{\tau_{dis}}$ exponentially converge to a smaller circular plane whose radius is $\lambda_{min}^{-1}(\boldsymbol{\Psi})\delta_{\tau_{dis}}$. The convergence rate and accuracy of estimation are simply improved by assigning larger eigenvalues to $\boldsymbol{\Psi}$, i.e., using higher bandwidth $\left(g_{DOb}\right)$ in the design of the observer. However, it cannot be freely increased due to practical constraints such as encoder resolution and sampling time. A DOb becomes more noise-sensitive as its bandwidth is increased.

As $\mathbf{e}(t) \rightarrow \mathbf{0}$, $\hat{\mathbf{z}}_1 \rightarrow \mathbf{z_1}$, $\hat{\mathbf{z}}_2 \rightarrow \mathbf{z_2}$ and $\hat{\mathbf{z}}_3 \rightarrow \mathbf{z_3}$. The estimations of disturbances and their successive derivatives are obtained by using Eq. (10-12).

Since force control dynamics has only matched disturbances (See Eq. (8)), conventional, i.e., zero-order, DOb can be used in the robust force controller synthesis [28].

## IV. Robust Motion Controller Synthesis

Sliding mode position and force controllers are designed for SEAs in this section. Continuous control signal is obtained by using two different methods, namely Quasi-SMC and Continuous-SMC. The stability of the proposed motion controllers is proved via Lyapunov's second method.

### A. Robust Position Controller Synthesis:

Eq. (1) can be rewritten by using

$$
\begin{aligned}
\dot{x}_1 &= x_2 \\
\dot{x}_2 &= \frac{k^n}{J_l^n} x_3 - d_2 \\
\dot{x}_3 &= x_4 \\
\dot{x}_4 &= \frac{1}{J_m^n} \tau_m - d_4
\end{aligned}
\tag{21}
$$

where $x_1 = q$; $x_2 = \dot{q}$; $x_3 = \theta$; $x_4 = \dot{\theta}$; $d_2 = \frac{k^n}{J_l^n} x_1 + \frac{b_l^n}{J_l^n} x_2 + \frac{\tau_l^d}{J_l^n}$ is the mismatched disturbance in the second-channel; and $d_4 = \frac{k^n}{J_m^n}(x_3 - x_1) + \frac{b_m^n}{J_m^n} x_4 + \frac{\tau_m^d}{J_m^n}$ is the matched disturbance in the control input channel, i.e., the fourth-channel.

The trajectory tracking error of the actuator's link and its successive time derivatives are:

$$
\begin{aligned}
e_P &= q^{des} - x_1 \\
\dot{e}_P &= \dot{q}^{des} - x_2 \\
\ddot{e}_P &= \ddot{q}^{des} - \frac{k^n}{J_l^n} x_3 + d_2 \\
\dddot{e}_P &= \dddot{q}^{des} - \frac{k^n}{J_l^n} x_4 + \dot{d}_2
\end{aligned}
\tag{22}
$$

where $q^{des}$ is the desired trajectory of the actuator's link, and $\dot{q}^{des}, \ddot{q}^{des}$ and $\dddot{q}^{des}$ are the successive time derivatives of $q^{des}$.

Let us describe a sliding variable in terms of the trajectory error and its successive time derivatives by using

$$
\sigma_P = \dddot{e}_P + c_{2_P} \ddot{e}_P + c_{1_P} \dot{e}_P + c_{0_P} e_P
\tag{23}
$$

where $c_{0_P}, c_{1_P}$ and $c_{2_P}$ are nonnegative real numbers [29, 30]. They can be tuned by using $c_{2_P} = 3 g_{SMC}, c_{1_P} = 3 g_{SMC}^2$ and $c_{0_P} = g_{SMC}^3$ where $g_{SMC} \in \mathbb{R}$ represents the bandwidth of the performance controller. As $g_{SMC}$ is increased the position control error converges to zero faster, yet the control system becomes more noise-sensitive [29]. It should be tuned by considering the practical constraints of the system, such as the resolutions of encoders.

The derivative of Eq. (23) is

$$
\dot{\sigma}_P = -\alpha_P \tau_m + \beta_P
\tag{24}
$$

where $\alpha_P = \frac{k^n}{J_m^n J_l^n}$; and $\beta_P = \dddot{q}^{des} + c_{2_P} \ddot{q}^{des} + c_{1_P} \ddot{q}^{des} + c_{0_P} \dot{q}^{des} - \frac{k^n}{J_l^n}(c_{0_P} x_2 + c_{1_P} x_3 + c_{2_P} x_4) + c_{1_P} d_2 + c_{2_P} \dot{d}_2 + \ddot{d}_2 + \frac{k^n}{J_l^n} d_4$.

The robust position control signal is generated as follows:

$$
\tau_m = \alpha_P^{-1} \rho_P \, \mathrm{sgn}(\sigma_P) + \alpha_P^{-1} \hat{\beta}_P
\tag{25}
$$

where $\rho_P > 0$ is the discontinuous control gain of SMC, $\mathrm{sgn}(\ )$ is the signum function, and

$$
\hat{\beta}_P = \dddot{q}^{des} + c_{2_P} \ddot{q}^{des} + c_{1_P} \ddot{q}^{des} + c_{0_P} \dot{q}^{des} - \frac{k^n}{J_l^n}(c_{0_P} x_2 + c_{1_P} x_3 + c_{2_P} x_4) + \\
c_{1_P} \hat{d}_2 + c_{2_P} \dot{\hat{d}}_2 + \ddot{\hat{d}}_2 + \frac{k^n}{J_l^n} \hat{d}_4
\tag{26}
$$

where $\hat{d}_2, \dot{\hat{d}}_2, \ddot{\hat{d}}_2$ and $\hat{d}_4$ are the estimations of $d_2, \dot{d}_2, \ddot{d}_2$ and $d_4$, respectively.

### B. Robust Force Controller Synthesis:

Eq. (8) can be rewritten by using

$$
\begin{aligned}
\dot{x}_1 &= x_2 \\
\dot{x}_2 &= \frac{1}{J_m^n} \tau_m - d
\end{aligned}
\tag{27}
$$

where $x_1 = \theta$; $x_2 = \dot{\theta}$; and $d = \frac{b_m^n}{J_m^n} \dot{\theta} + \frac{\tilde{\tau}_m^d}{J_m^n}$ is the matched disturbance in force/impedance control.

The trajectory tracking error of the SEA's motor, i.e., the trajectory tracking error of the actuator's spring force/torque, and its derivative are

$$
\begin{aligned}
e_F &= \theta^{des} - x_1 \\
\dot{e}_F &= \dot{\theta}^{des} - x_2
\end{aligned}
\tag{28}
$$

where $\theta^{des}$ is given in Eq. (5), and $\dot{\theta}^{des}$ is the derivative of $\theta^{des}$.

Let us define a sliding variable in terms of the trajectory error and its time derivative by using Eq. (28).

$$
\sigma_F = \dot{e}_F + c_{0_F} e_F
\tag{29}
$$

where $c_{0_F}$ is a nonnegative real number. The convergence rate of the force control error can be similarly adjusted by tuning $c_{0_F}$, e.g., as $c_{0_F}$ is increased the force control error goes to zero faster yet the noise-sensitivity deteriorates.

The derivative of Eq. (29) is

$$
\dot{\sigma}_F = -\alpha_F \tau_m + \beta_F
\tag{30}
$$

where $\alpha_F = 1/J_m^n$, and $\beta_F = \ddot{\theta}^{des} + c_{0_F} \dot{\theta}^{des} - c_{0_F} x_2 + d$.

The robust force control signal is generated as follows:

$$
\tau_m = \alpha_F^{-1} \rho_F \, \mathrm{sgn}(\sigma_F) + \alpha_F^{-1} \hat{\beta}_F
\tag{31}
$$

where $\rho_F > 0$ is the discontinuous control gain of SMC, and

$$
\hat{\beta}_F = \ddot{\theta}^{des} + c_{0_F} \dot{\theta}^{des} - c_{0_F} x_2 + \hat{d}
\tag{32}
$$

where $\hat{d}$ is the estimation of $d$.

It is noted that Eq. (31) requires the acceleration measurement of the SEA's link [27]. The control signal can be generated without acceleration measurement by using

$$
\hat{\beta}_F = c_{0_F} \dot{\theta}^{des} - c_{0_F} x_2 + \hat{d}
\tag{33}
$$

The block diagram of the proposed motion controller is illustrated in Fig. 2.

### C. Stability Analysis of the Robust Motion Controllers:

Let us consider the following Lyapunov function candidate.

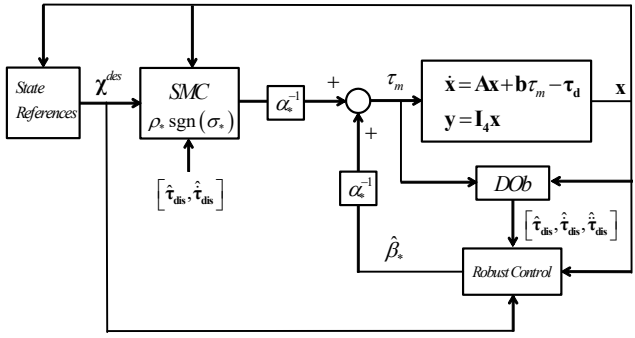

Fig. 2: Block diagram of the robust motion controller. * is P in position control and F in force control. $\chi^{des}$ represents reference trajectories, i.e., it is $[q^{des}, \dot{q}^{des}, \ddot{q}^{des}, \dddot{q}^{des}]$ in position control and $[\theta^{des}, \dot{\theta}^{des}, \ddot{\theta}^{des}]$ in force control.

$$V = \frac{1}{2}\sigma_*^2 \qquad (34)$$

where $*$ is $P$ in position control and $F$ in force control.

The derivative of Eq. (34) is

$$\dot{V} = \sigma_* \left( -\rho_* \operatorname{sgn}(\sigma_*) - \left( \beta_* - \hat{\beta}_* \right) \right) \qquad (35)$$

$$\dot{V} \le -\left( \rho_* - \left| \beta_* - \hat{\beta}_* \right| \right) |\sigma_*| \qquad (36)$$

Since disturbance estimation error is bounded when the observer is properly tuned, $\beta_*$ is also bounded. It can be expressed as follows:

$$\left| \beta_P - \hat{\beta}_P \right| \le c_{1_P} \left| d_2 - \hat{d}_2 \right| + c_{2_P} \left| \dot{d}_2 - \dot{\hat{d}}_2 \right| + \left| \ddot{d}_2 - \ddot{\hat{d}}_2 \right| + \frac{k^n}{J_l^n} \left| d_4 - \hat{d}_4 \right| \le \delta_{\beta_P}$$

$$\left| \beta_F - \hat{\beta}_F \right| \le \left| \dot{\theta}^{des} \right| + \left| d - \hat{d} \right| \le \delta_{\beta_F} \qquad (37)$$

where $\delta_{\beta_P} > 0$ and $\delta_{\beta_F} > 0$ are real numbers.

If SMC gain is $\rho_* = \delta_{\beta_*} + \dfrac{\mu}{\sqrt{2}}$ where $\mu > 0$, then $\sigma_*(t)$ converges to zero in a finite time, i.e.

$$T \le \frac{\sqrt{2}}{\mu} |\sigma_*(t_0)| \qquad (38)$$

where $T$ is the maximum convergence time, and $t_0$ is the arbitrary initial time. Hence, the trajectory tracking errors given in Eq. (22) and (28) exponentially go to zero [29, 30].

Compared to the robust position control problem of an SEA, its robust force control problem suffers from only the matched disturbances. Although the matched disturbances can be directly suppressed by using conventional SMC, the force controller may suffer from chattering due to high SMC gain in practice. In the proposed robust motion controllers, DOb not only improves the robustness against disturbances but also suppresses the chattering by allowing to lower the discontinuous SMC gain constraint.

### D. Continuous Robust Motion Controller Synthesis:

#### Quasi-SMC:

Continuous control signal can be obtained by using the approximation of the signum function as follows:

$$\operatorname{sgn}(\sigma_*) \cong \frac{\sigma_*}{|\sigma_*| + \varepsilon} \qquad (39)$$

where $\varepsilon \ne 0 \in \mathbb{R}$.

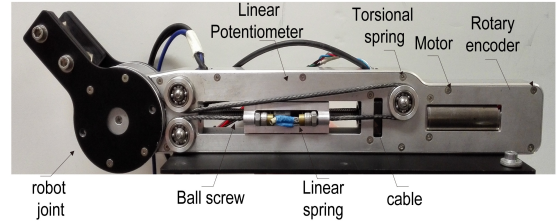

Fig. 3: A prototype of the novel series elastic actuator.

| Parameters | Description | Values |
|---|---|---|
| $J_m^n$ | Inertia of motor | $2.2 \times 10^{-6}\ kgm^2$ |
| $J_l^n$ | Inertia of link | $4 \times 10^{-6}\ kgm^2$ |
| $k^n$ | Torsional spring stiffness | $0.14\ Nm/rad$ |

There is a trade-off between the control signal chattering and the robustness against disturbances in Quasi-SMC. As $\varepsilon$ is increased, the fluctuation of the control signal is suppressed yet the robustness deteriorates. However, as $\varepsilon$ is decreased, the robustness improves yet the fluctuation of the control signal is increased.

#### Continuous-SMC:

Without degrading the robustness of the proposed motion controllers, continuous control signal can be obtained by using Continuous-SMC.

Let us design new sliding variables by using

$$\tilde{\sigma}_P = \dddot{e}_P + \tilde{c}_{3_P} \ddot{e}_P + \tilde{c}_{2_P} \ddot{e}_P + \tilde{c}_{1_P} \dot{e}_P + \tilde{c}_{0_P} e_P \qquad (40)$$
$$\tilde{\sigma}_F = \ddot{e}_F + \tilde{c}_{1_F} \dot{e}_F + \tilde{c}_{0_F} e_F$$

where $\tilde{c}_{0_P}, \tilde{c}_{1_P}, \tilde{c}_{2_P}, \tilde{c}_{3_P}, \tilde{c}_{0_F}$ and $\tilde{c}_{1_F}$ are positive real numbers, and they can be similarly tuned to adjust the convergence rates of the position and force control errors.

Continuous control signal, which makes sliding variables $(\tilde{\sigma}_P \text{ and } \tilde{\sigma}_F)$ to converge zero in a finite time, is designed as follows [30]:

$$\tau_m = \int \alpha_*^{-1} \rho_* \operatorname{sgn}(\tilde{\sigma}_*) dt + \alpha_*^{-1} \hat{\beta}_* \qquad (41)$$

where $*$ is $P$ in position control and $F$ in force control.

Continuous control signal is obtained by integrating the signum function in Eq. (41). The main drawback of Continuous-SMC is that the acceleration measurement is required in the design of the sliding variables in Eq. (40).

### V. EXPERIMENTS

This section verifies the proposed robust motion controller with the experimental results of a novel SEA illustrated in Fig. 3. The novel SEA has a two-state variable-stiffness actuation mechanism which consists of serially connected a hard torsional spring and a soft linear spring [25]. The position and force control experiments were conducted by neglecting the linear spring of the novel SEA; i.e., a conventional SEA structure with a torsional spring was used in the experiments. The nominal parameters of the SEA's dynamic model are given in Table I. To implement the sliding mode motion controllers, a dSPACE real-time control system was used by setting its sampling frequency at 2 KHz. The resolutions of the motor (Maxon EC-4pole) and link encoders were 2048 PPR and 1024 PPR, respectively.

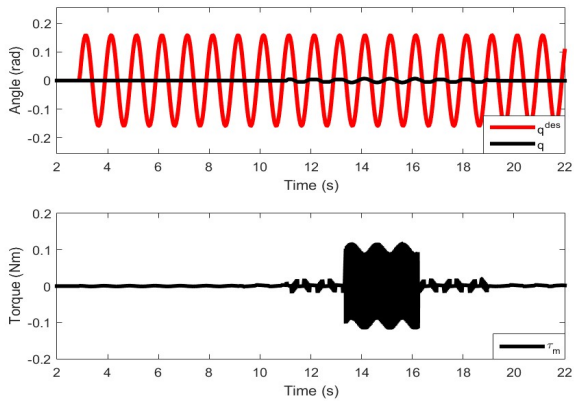

a) Trajectory tracking control experiment via conventional sliding mode position controller. Reference is a sine wave with 0.1592 rad amplitude and 1 Hz frequency.

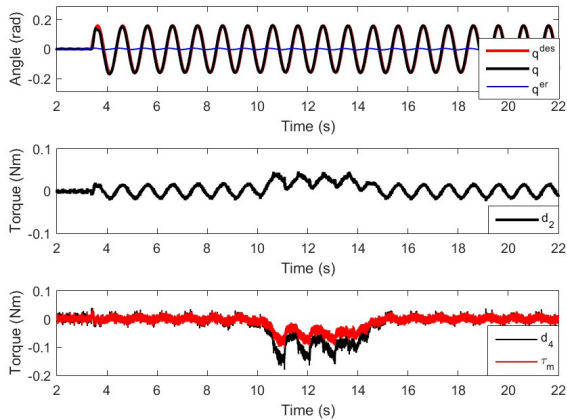

b) Trajectory tracking control experiment via the proposed sliding mode position controller. Reference is a sine wave with 0.1592 rad amplitude and 1 Hz frequency. $q^{er} = q^{des} - q$.

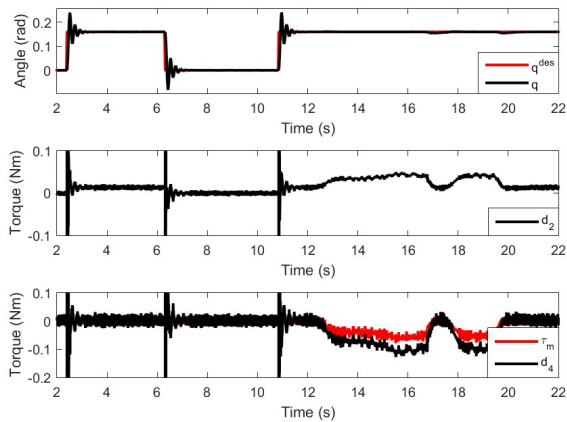

c) Regulation control experiment via the proposed sliding mode position controller. Reference is a step input with 0.1592 rad magnitude.

Fig. 4: Position control experiments when discontinuous SMC-based controllers are implemented. $\rho_p = 0.001$ and $g_{DOb} = 500$ rad/s .

Let us begin by presenting position control experiments with discontinuous SMC method. Fig. 4a illustrates a trajectory tracking experiment when conventional sliding mode controller is implemented; i.e., the mismatched disturbances are neglected in the sliding mode controller synthesis. A constant weight was attached to the link of the actuator that did not interact with environment. To improve the robustness, the discontinuous control gain was gradually increased between 9 and 14 seconds. It is clear from the figure that the SEA's link cannot track the reference sine wave by using a conventional sliding mode controller due to disturbances. As the discontinuous control gain was increased, the link of the actuator started to follow the sinusoidal reference input. However, the performance of the trajectory tracking control was very poor, and the chattering of the control signal was very high. It is illustrated between 9 and 19 seconds in this figure. Fig. 4b illustrates a trajectory tracking experiment when the proposed sliding mode controller is implemented; i.e., the sliding surface is designed by using the estimations of the disturbances. The same weight was attached to the link of the actuator, and it was disturbed with random external disturbances between 10 and 15 seconds as shown in the figure (see the estimations of the matched and mismatched disturbances). This figure shows that the link of the SEA can track the sine wave with high-accuracy when the disturbances are suppressed by using the proposed sliding mode controller. A regulation control experiment is illustrated in Fig. 4c. Similarly, the link of the SEA was disturbed with random external disturbances between 13 and 19 seconds (see the estimations of the matched and mismatched disturbances). As shown in the figure, the proposed controller can precisely suppresses disturbances, and the actuator's link can track step input with high accuracy. The transient response of the SEA's link suffered from vibration as the resonance modes of the compliant actuator were neglected in the sliding mode controller synthesis [9]. The vibration of the SEA's link can be suppressed by either using a continuous trajectory as shown in Fig. 4b or implementing a damping controller. Fig. 4b and Fig. 4c show that the discontinuous SMC-based robust position controller does not suffer from high control signal chattering since lower SMC gains can be used by cancelling disturbances via DOb.

Let us now present position control experiments with continuous SMC methods, namely Quasi-SMC and Continuous-SMC. All the experiments were conducted by using the proposed sliding mode controller which treated the mismatched disturbances in the controller synthesis. The link of the SEA was similarly disturbed with random external disturbances between 7 and 14 seconds as shown in Fig. 5. Fig. 5a and Fig. 5b illustrate trajectory tracking experiments when Quasi-SMC is implemented with different approximations of the signum function. As $\varepsilon$ was increased, the sliding mode controller became more sensitive to disturbances, and the accuracy of the link trajectory deteriorated. The control signal fluctuations were similar when different $\varepsilon$ values were used in Quasi-SMC because low discontinuous control gains were applied by cancelling disturbances via DOb in the proposed sliding mode position controller. Fig. 5c illustrates trajectory tracking experiment when Continuous-SMC is implemented. The position control performance of the Continuous-SMC-based robust controller was better than that of the Quasi-SMC-based robust controller. The link of the SEA tracked the sine wave with high-accuracy. However, the Continuous-SMC-based robust controller suffered from noise because acceleration

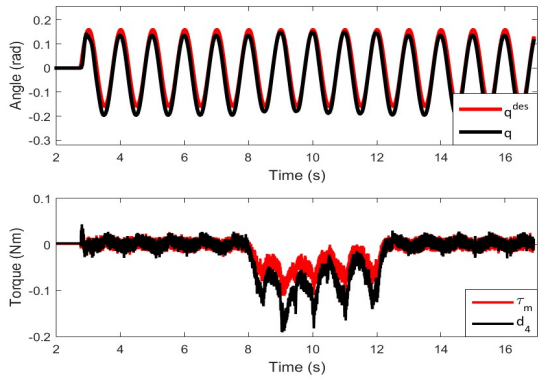

a) Trajectory tracking control experiment via the proposed sliding mode position controller with Quasi-SMC. $\varepsilon = 0.1$.

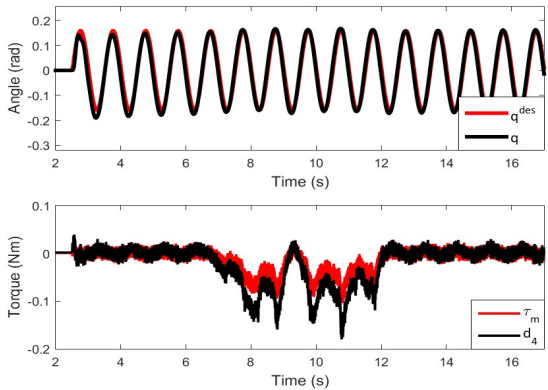

b) Trajectory tracking control experiment via the proposed sliding mode position controller with Quasi-SMC. $\varepsilon = 0.001$.

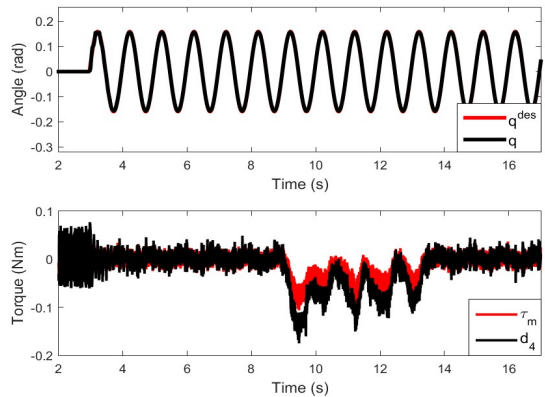

c) Trajectory tracking control experiments via the proposed sliding mode position controller with Continuous-SMC.

Fig. 5: Position control experiments when continuous SMC-based controllers are implemented. Reference is a sine wave with 0.1592 rad amplitude and 1 Hz frequency. $\rho_p = 0.001$ and $g_{DOb} = 500$ rad/s.

estimation, which was obtained by differentiating position measurement twice, was used in the controller synthesis.

Fig. 4b and Fig. 5 show that control signal chattering is not a severe problem in the proposed sliding mode controller. Since similar tracking performance can be achieved with practical control signal, authors recommend discontinuous SMC method for SEAs.

Last, let us present force/impedance control experiments with discontinuous SMC method. Active and passive robust force/impedance control experiments were performed to validate the proposed sliding mode force controller. In the former, force trajectory tracking and regulation control experiments were performed by contacting with a passive stiff environment (metal). In the latter, a physical human-robot interaction experiment was performed when zero-force control was implemented. Fig. 6a illustrates a force/impedance trajectory tracking experiment when the SEA's link initially contacts with the stiff environment. It is clear from the figure that the trajectory of the spring force is tracked with high-accuracy by suppressing the matched disturbances. Since the matched disturbances were cancelled by using DOb in the inner-loop, the robustness of the force control system was achieved without suffering from control

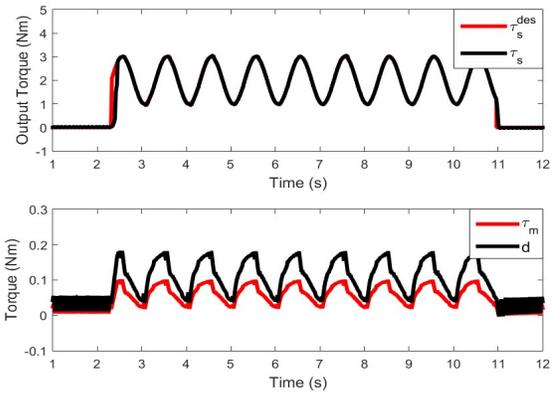

a) Trajectory tracking experiment. The reference is $2 + \sin(2\pi t)$ Nm.

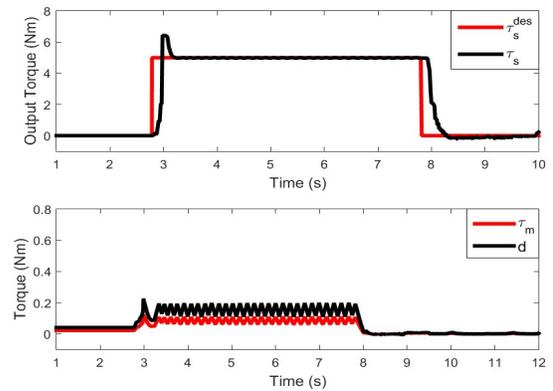

b) Regulation control experiment. Reference is 5 Nm step input.

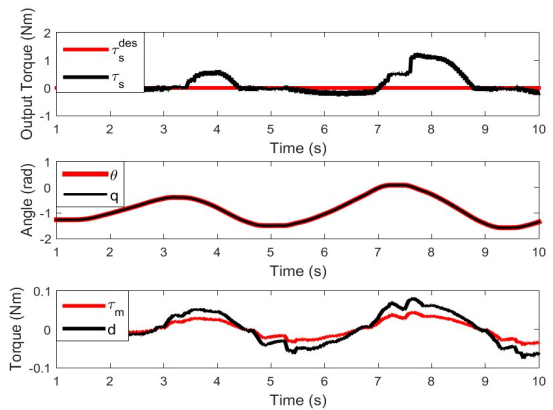

c) Physical human-robot interaction experiment.

Fig. 6: Force control experiments via the proposed sliding mode force controller with discontinuous SMC. $\rho_F = 0.0035$ and $g_{DOb} = 500$ rad/s.

signal chattering. Fig. 6b illustrates a force/impedance regulation experiment when the SEA's link does not initially contact with the stiff environment. Since the approaching speed of the SEA's link was neglected in this experiment, a high-overshoot was observed in the beginning of the contact motion. To eliminate or lower the overshoot in force control, one should treat approaching speed of the actuator's link [25]. The figure shows that the regulation experiment can be conducted with high-accuracy by using the proposed sliding mode force controller. Compared to the position regulation experiment, the performance of the force regulation experiment did not suffer from vibration. However, the resonance modes influenced the system as a disturbance [9]. The control signal was automatically adjusted so that the disturbance, i.e., the vibration, was precisely suppressed. The results of the physical human-robot interaction experiment are illustrated in Fig. 6c. Passive force control experiment was performed by pushing/pulling the SEA's link and applying zero-force reference input. As it is shown in the figure, the SEA can physically interact with a human being in a safe and sound manner when the proposed sliding mode force controller is implemented.

## VI. CONCLUSION

This paper has proposed a robust motion controller for SEAs by using DOb and SMC. All disturbances are lumped into fictitious disturbance variables so that practical dynamic models are derived for position and force control applications. It is shown that the force control dynamics is of second-order and suffers from matched disturbances; however, the position control dynamics is of fourth-order and suffers from mismatched disturbances as well as matched ones. Therefore, an SEA's position control problem is more complicated than its force control problem. By combining DOb and SMC, not only the disturbances of an SEA are precisely suppressed but also the discontinuous control signal chattering is significantly lowered. When the proposed sliding mode controllers are implemented, 1) the position and force trajectories of an SEA can be tracked with high-accuracy; 2) and an SEA can physically interact with active and passive environments in a safe and sound manner. Position and force control experiments are presented to validate the robust motion controllers.